\documentclass[runningheads]{llncs}
\usepackage{graphicx}
\usepackage{amsmath,amssymb} 
\usepackage{color}
\usepackage{threeparttable}
\usepackage{multirow}
\usepackage{array}
\usepackage{subfig}
\usepackage[noend]{algpseudocode}
\usepackage{algorithmicx,algorithm}
\usepackage{booktabs}    
\usepackage[breaklinks=true,bookmarks=false]{hyperref}
\hypersetup{
    colorlinks=true,
    citecolor=darkblue
}
\definecolor{darkblue}{rgb}{0,0.08,0.45}
\usepackage[width=122mm,left=12mm,paperwidth=146mm,height=193mm,top=12mm,paperheight=217mm]{geometry}

\begin{document}
\mainmatter

\def\ECCV18SubNumber{***}  
\title{Discretization-Aware Architecture Search}
\author{Yunjie Tian\inst{1},
        Chang Liu\inst{1},
        Lingxi Xie\inst{2},
        Jianbin Jiao\inst{1},
        Qixiang Ye\inst{1}}

\institute{University of Chinese Academy of Sciences, \and
Noah’s Ark Lab, Huawei}

\maketitle

\begin{abstract}
The search cost of neural architecture search (NAS) has been largely reduced by weight-sharing methods. These methods optimize a super-network with all possible edges and operations, and determine the optimal sub-network by discretization, \textit{i.e.}, pruning off weak candidates. The discretization process, performed on either operations or edges, incurs significant inaccuracy and thus the quality of the final architecture is not guaranteed. This paper presents discretization-aware architecture search (DA\textsuperscript{2}S), with the core idea being adding a loss term to push the super-network towards the configuration of desired topology, so that the accuracy loss brought by discretization is largely alleviated. Experiments on standard image classification benchmarks demonstrate the superiority of our approach, in particular, under imbalanced target network configurations that were not studied before. The code is available at \href{https://github.com/sunsmarterjie/DAAS}{\color{magenta}https://github.com/sunsmarterjie/DAAS}.

\end{abstract}

\section{Introduction}
 
Network architecture search (NAS) is a research topic aimming to explore the design of neural networks in a large space that is not well covered by human expertise. To alleviate the computational burden of the reinforcement-based~\cite{zoph2016neural,zoph2018learning} and evolutionary~\cite{real2017large,xie2017genetic,real2018regularized} algorithms that evaluate sampled architecture individually, researchers proposed one-shot search methods~\cite{brock2017smash} which first optimized a super-network with all possible architectures included, and then sampled sub-networks from it for evaluation~\cite{pham2018efficient}. By sharing computation, this kind of methods accelerated NAS by $3$--$4$ orders of magnitudes.

A representative example of one-shot search is differentiable architecture search (DARTS~\cite{liu2018darts}), which formulates the super-network into a differentiable form with respect to a set of architectural parameters, \textit{e.g.}, operations and connections, so that the entire NAS process can be optimized in an end-to-end manner. DARTS did not require an explicit process for evaluating each sub-network, but performed a standalone discretization process to determine the optimal sub-architecture, on which re-training is performed. Such an efficient search strategy does not require the search cost to increase dramatically as the size of search space, and the space can be much larger compared with other NAS approaches.

\begin{figure}[t]
\centering
\includegraphics[width=1.0\columnwidth]{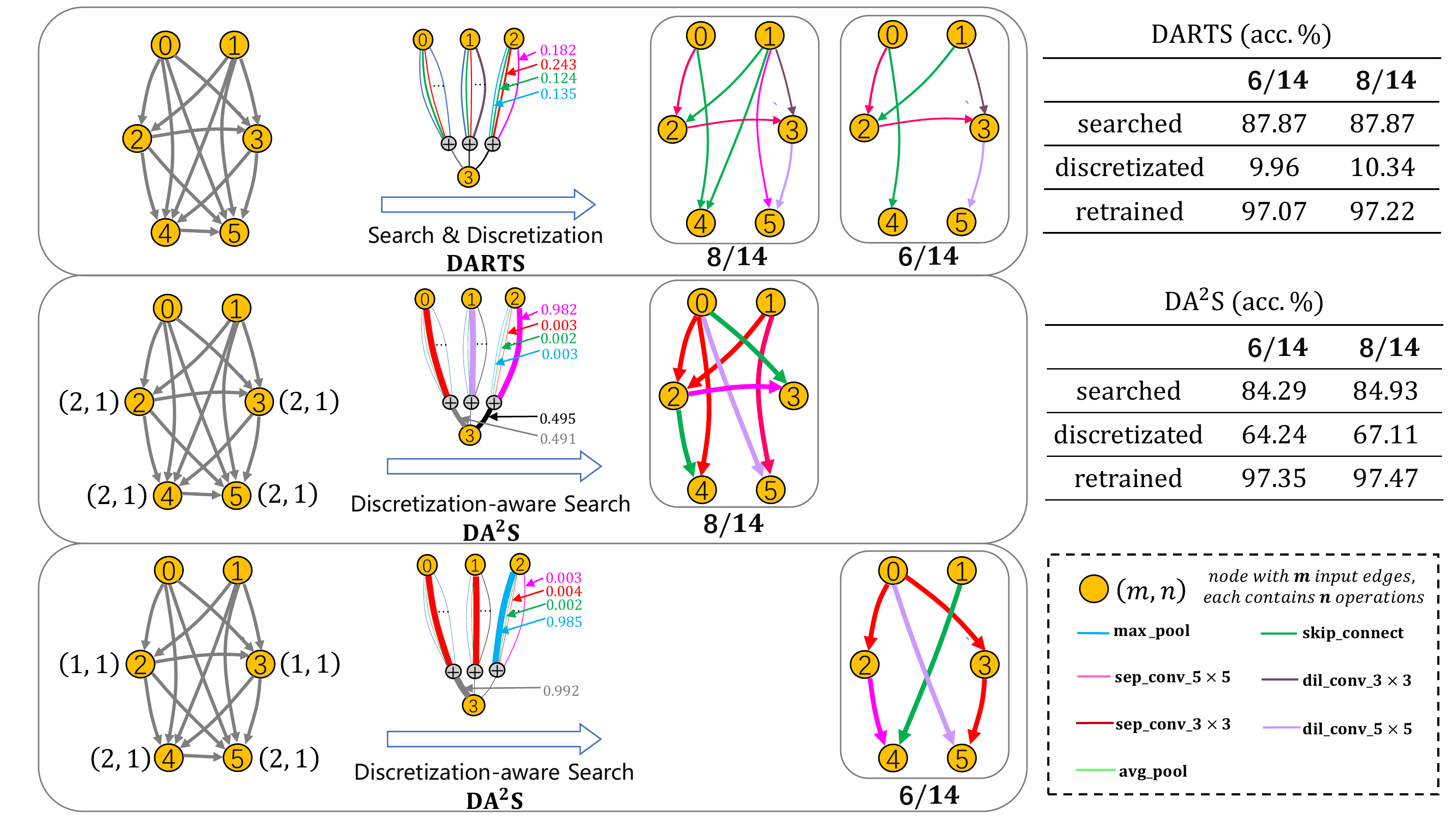}
\caption{\textbf{Top}: the normal cell of DARTS, from which we investigate node3 which sums up $3$ input edges in the search stage. In two discretization configurations (preserving $8$ and $6$ out of $14$ edges), this node can preserve $2$ and $1$ input(s), respectively, but pruning off inputs with moderate weights can lead to dramatic super-network accuracy and unsatisfying re-training accuracy. \textbf{Middle \& Bottom}: DA\textsuperscript{2}S is aware of the number of inputs to be preserved for each node and pushes weights to get close to either 1 or 0, so as the discretization loss is largely alleviated, and re-training accuracy improved. This figure is best viewed in color.}
\label{fig:motivation}
\end{figure}

Despite of the superiority about efficiency, DARTS is believed to suffer the gap between the optimized super-network and the sampled sub-networks. In particular, as illustrated in~\cite{chen2019progressive}, the difference between the number of cells can cause a `depth gap', and the search performance is largely stabilized by alleviating the gap. In this paper, we point out another gap, potentially more important, caused by the process of discretizing architectural weights of the super-network. To be specific, DARTS combines candidate operations and edges with a weighted sum (the weights are learnable), and preserves a fixed number of candidates with strong weights and discards others. 
However, there is no guarantee that the discarded weights are relatively small -- if not, this discretization process can introduce significant inaccuracy in neural responses to \textbf{each} cell.
Such inaccuracy accumulates and finally causes that \textbf{a well-optimized super-network does not necessarily generates high-quality sub-networks},
in particular (i) when the the discarded candidates still have moderate weights; and/or (ii) the number of pruned edges is relatively small compared to that in the super-network. 
Figure~\ref{fig:motivation} shows a cell optimized by DARTS. One can see that discretization causes the super-network accuracy to drop dramatically, which also harms the performance of searched architecture in the re-training stage.

To alleviate the above issue, we propose discretization-aware architecture search (DA\textsuperscript{2}S). The main idea is to introduce an additional term to the loss function, so that the architectural parameters of the super-network is gradually pushed towards the desired configuration during the search process. To be specific, we formulate the new loss term into an entropy function based on the property that minimizing the entropy of a system drives maximizing the sparsity and discretization of the elements (weights) in the system. The objective of entropy is to enforce each weight to get close to either $0$ or $1$, with the number of $1$'s determined by the desired configuration, so that the discretization process, by removing candidates with weights close to $0$, does not incur significant accuracy loss. Being differentiable to architectural parameters, the entropy function can be freely plugged into the system for SGD optimization.
We perform experiments on two standard image classification benchmarks, namely, CIFAR10 and ImageNet, based on PC-DARTS~\cite{xu2020pcdarts}, an efficient variant of DARTS. Note that two sets of architectural parameters exist in PC-DARTS, taking control of operations in an edge and edges that sum into a node, respectively, and they are potentially equipped with different loss terms. We evaluate different configurations (\textit{i.e.}, varying from each other in the number of preserved edges for each node), most of which have not been studied before. When each search process reaches the end, the super-network converges into a discretization-friendly form, and the discretization process causes much smaller accuracy drop than that reported without the entropy loss. Consequently, the searched architecture, under any configuration, enjoys higher yet more stable performance, and the advantage is more significant as the configuration becomes more imbalanced, on which the original search method suffers a larger `discretization gap'.

\section{Related Work}
\label{RelatedWork}
The rapid development of deep learning~\cite{lecun2015deep}, in particular convolutional neural networks, have largely changed the way of designing computational models in computer vision. Recent years have witnessed a trend of stacking more and more convolutional layers to a deep network~\cite{krizhevsky2012imagenet,simonyan2014very,he2016deep,huang2017densely} so that more trainable parameters are included and higher recognition accuracy is achieved. 

Going one step further, researchers started to consider the possibility that designs deep networks automatically, and thereby created a new research area termed neural architecture search (NAS)~\cite{zoph2016neural}. NAS defines a sub-field of automated machine learning (AutoML)~\cite{AutoWEKA} and has attracted increasing attentions in both academia and industry. The idea is to construct a sufficiently large space and thus enables the architecture to adjust according to training data, simulating the process of evolutionary computation. With carefully monitored search strategies, NAS has claimed better performance compared to hand-designed networks in a wide range of applications including image classification~\cite{zoph2016neural}, object detection~\cite{nasfpn}, and semantic segmentation~\cite{liu2019auto}.

The early efforts of NAS mainly involved heuristic search in a very large space, and the sampled architectures were often evaluated individually. Representative examples include using reinforcement learning (RL) to formulate network or block designs~\cite{zoph2016neural,zoph2018learning,liu2018progressive}, applying evolutionary algorithms (EA) to force the network evolve throughout iterations~\cite{real2017large,xie2017genetic}, or simply performing guided random search to find competitive solutions~\cite{real2018regularized}. These methods often require a vast amount of computation, \textit{e.g.}, thousands of GPU-days. To accelerate the search process, one-shot architecture search was proposed to share computation among architectures with similar building blocks~\cite{brock2017smash}.

One-shot architecture search was later developed into weight-reusing~\cite{cai2018efficient} and weight-sharing~\cite{pham2018efficient} methods which can reduce the search costs by orders of magnitudes. Beyond this point, researchers proposed to improve the search stability using better sampling methods~\cite{singlepath,fairnas}, explored the importance of the search space~\cite{mobilenetv3}, and tried to integrate hardware consumption such as latency as additional evaluation metrics~\cite{tan2018mnasnet}. These efforts eventually leads to powerful architectures that achieve state-of-the-art performance on ImageNet~\cite{cai2018efficient} with moderate computational cost overhead.

A special family of one-shot architecture search falls into formulating the search space into a super-network which can adjust itself in a continuous space~\cite{luo2018neural}. Based on this, the network and architectural parameters can be jointly optimized, which leads to a differentiable approach for architecture search. DARTS~\cite{liu2018darts}, a representative differentiable framework, designed an over-parameterized super-network which contains exponentially many sub-networks with shared weights. It performed bi-level optimization to update network weights and architectural weights alternately and, at the end of the search stage, used a greedy algorithm to prune off the operations and edges with lower weights. Partially-Connected DARTS~\cite{xu2020pcdarts} pursed a more efficient search by sampling a small part of super-network to reduce the redundancy in exploring the network space.

Recent DARTS methods~\cite{chen2019progressive,xu2020pcdarts} have achieved success on both architecture quality and search efficiency. Nevertheless few researchers noticed that the discretization process incurs a significant accuracy loss, which makes it difficult to obtain a high-quality sub-network from the optimal sub-network~\cite{zela2020understanding}. This paper investigates this problem born with DARTS methods in a systematic way with the target to search discretization-aware architectures from the perspective of model regularization.

\section{Discretization-Aware Architecture Search}
\label{sec:approach}

\subsection{Preliminaries: DARTS}

DARTS~\cite{liu2018darts} designs a cell-based search space to facilitate efficient differentiable architecture search. Each cell is represented as a directed acyclic graph with $N$ nodes, where each node defines a network layer. There is a pre-defined space of operations denoted by $\mathcal{O}$, where each element, $o\!\left(\cdot\right)$, denotes a fixed operation. Commonly used operations include identity connection, and $3\times3$ convolution performed at a network layer.

Within a cell, the searching goal is to choose one operation from $\mathcal{O}$ for each pair of nodes. Let $\left(i,j\right)$ denote a pair of nodes, where ${0}\leqslant{i}<{j}\leqslant{N-1}$. The primary idea of DARTS is to formulate the information and gradient propagated from $i$ to $j$ as a weighted sum over $\left|\mathcal{O}\right|$ operations, as 
${f_{i,j}\!\left(\mathbf{z}_i\right)}={{\sum_{o\in\mathcal{O}}}a^{o}_{i,j}\cdot o\!\left(\mathbf{z}_i\right)}$,
where $a^{o}_{i,j}=\frac{\exp\left\{\alpha_{i,j}^o\right\}}{{\sum_{o'\in\mathcal{O}}}\exp\left\{\alpha_{i,j}^{o'}\right\}}$ and $\mathbf{z}_i$ denotes the output of the $i$-th node, and $\boldsymbol{\alpha}$ is a set of architectural parameters to weight operations within each edge, with $\alpha_{i,j}^o$ determining the weight of $o\!\left(\cdot\right)$ in edge $\left(i,j\right)$. 
Following PC-DARTS~\cite{xu2020pcdarts}, we introduce an extra set of architectural parameters ($\boldsymbol{\beta}$) in our DA\textsuperscript{2}S to determine the weight of each edge. Thus, the output of a node is the sum of all input flows, \textit{i.e.},
${\mathbf{z}_j}={{\sum_{i<j}}b_{i,j}\cdot f_{i,j}\!\left(\mathbf{z}_i\right)}$, where $b_{i,j}=\frac{\exp\left\{\beta_{i,j}\right\}}{{\sum_{i'<j}}\exp\left\{\beta_{i',j}\right\}}$.
The output of the entire cell is formed by concatenating the output of all prior nodes, \textit{i.e.}, $\mathrm{concat}\!\left(\mathbf{z}_2,\mathbf{z}_3,\ldots,\mathbf{z}_{N-1}\right)$. Note that the first two nodes, $\mathbf{z}_0$ and $\mathbf{z}_1$, are input nodes to a cell, which are fixed during the search procedure.

This design makes the entire framework differentiable to both layer weights and hyper-parameters $\alpha_{i,j}^o$, so that it is possible to perform architecture search in an end-to-end fashion. After the search process is completed, on each edge $\left(i,j\right)$, the operation $o$ with the largest $\alpha_{i,j}^o$ value is preserved, and each node $j$ is connected to two precedents ${i}<{j}$ with the largest $\alpha_{i,j}^o$ preserved. Denote the architectural parameters as $\boldsymbol{\alpha}=\{\alpha_{i,j}^o\}$, and the overall super-network as $F_{\boldsymbol{\alpha},\boldsymbol{\theta}}(\cdot)$, which is parameterized by both $\boldsymbol{\alpha}$ and $\boldsymbol{\theta}(\boldsymbol{\alpha})$. The learning procedure of DARTS optimizes the image classification loss to determine $\boldsymbol{\alpha}$ and $\boldsymbol{\theta}$, as
\begin{equation}
\label{eq:darts_loss}
\arg\min_{\boldsymbol{\alpha},\boldsymbol{\theta}} \mathcal{L}^\mathrm{C}(\boldsymbol{\alpha},\boldsymbol{\theta}) = \frac{1}{M}\sum_{m=1}^M\mathbf{y}_m^\top\cdot\log\big(F_{\boldsymbol{\alpha},\boldsymbol{\theta}}(\mathbf{x}_m)\big),
\end{equation}
where $\{\mathbf{x}_m,\mathbf{y}_m\}_{m=1}^M$ denotes a batch of training samples with corresponding class labels.

\subsection{The Devil is in the Discretization Loss}

It is well acknowledged that DARTS-based approaches suffer limited stability, \textit{i.e.}, when the same search procedure runs for several times individually, the searched architectures can report varying performance during the re-training stage. For this reason, the original DARTS~\cite{liu2018darts} evaluated the architectures found in four individual search phases on the validation set and picked up the best one, which results in $4\times$ search cost. More importantly, as the search space gets enlarged, the number of trials require to find a high-quality architecture may also increase, and finally, the DARTS-based approaches may lose the advantage in efficiency.

An important insight that our work delivers is that the instability is partly caused by the discretization loss. Here, by discretization we mean the process that picks up the best operation and/or edge and discards others according to the architectural weights of the super-network, \textit{i.e.}, the continuous parameters, $\boldsymbol{\alpha}$ and $\boldsymbol{\beta}$, are discretized so that a pre-defined number of elements are optimized towards 1 and others close to 0. This obviously introduces inaccuracy to the well-trained super-network. To show this, we follow DARTS to train a super-network on CIFAR10, which reports an accuracy of $87.87\%$ on the validation set. Then, we investigate the impact of discretization by replacing the corresponding part with the trained weights, \textit{e.g.}, on each edge, keeping the dominating operation (using a weight of $1$) with its parameters (\textit{e.g.}, convolutional weights) unchanged. Results are shown in Figure~\ref{fig:motivation}. The accuracy drop is dramatic, \textit{e.g.}, under the setting of DARTS (each node has $2$ edges preserved), the validation accuracy drops from $87.87\%$ to $10.34\%$. If we investigate a more imbalanced discretization (the first two nodes have $1$ edge each and the last two nodes have $2$ edges each), the validation accuracy drops to $9.96\%$, which is even close to a random guess. This is unexpected and violates the design nature of one-shot NAS, which suggests that dramatically bad sub-networks can be sampled from a well-trained super-network. Consequently, there is no guarantee that architectures found in this way can eventually report good performance, even after a complete re-training process has been performed.

We argue that such gap is caused by that the training process is not aware of that a discretization process will be performed afterwards. For example, when $\left|\mathcal{O}\right|$ operations are competing in an edge $\left(i,j\right)$, they `assume' that the input, $\mathbf{x}_i$ is a weighted sum of the outputs of all nodes prior to $i$. When discretization is performed, $\mathbf{x}_i$ is modified into the output of the dominating node, but the weights on edge $\left(i,j\right)$ may not match the new input. Such inaccuracy accumulates throughout the entire network and eventually leads to catastrophic accuracy drop. Therefore, the key to alleviate the gap is to make the search process aware of discretization, as well as the topology of the final architecture. We will elaborate our solution in the next part.

\subsection{Entropy-based Discretization-Aware Search}

\begin{figure}[t]
\centering
\includegraphics[width=1.0\columnwidth]{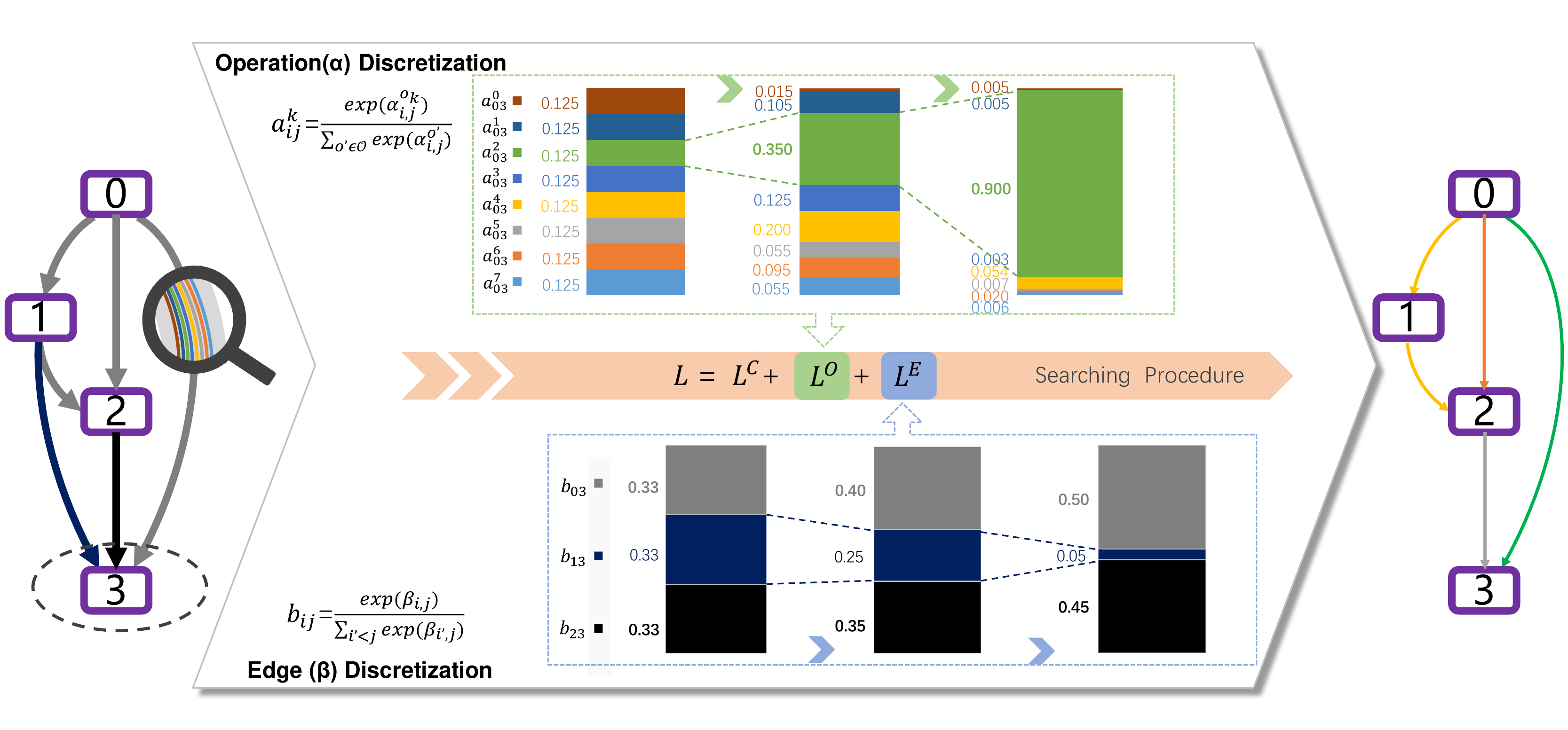}
\caption{Illustration of DA\textsuperscript{2}S which forces the softmax of architectural parameters $\alpha$ and $\beta$ moving towards extreme points. Here, $a_{i,j}^{k}$ denotes the weight on the $k$-th operation, and {$b_{i,j}$} denotes the weight between nodes $i$ and $j$. Each color indicates a candidate, and the area of each region corresponds to the weight of corresponding candidate. This figure is best viewed in color.}
\label{fig:pipeline}
\end{figure}

Figure~\ref{fig:pipeline} shows the overall framework of discretization-aware search. The main idea is to use the topology constraint to guide the optimization process, so that super-network eventually gets close to a sub-network that is allowed to appear as the final architecture. This is achieved by adding a loss function that measures the minimal distance between the current super-network and any acceptable sub-network. Specifically, we introduce an entropy-based loss function for each set of architectural parameters to fulfil this goal. 

Below we elaborate the details when applying this methodology to two sets of parameters, $\boldsymbol{\alpha}$ (operation) and $\boldsymbol{\beta}$ (edge), followed by discussions on the priority of discretization and the relationship between prior works and our DA\textsuperscript{2}S.

\vspace{0.2cm}
\noindent$\bullet$\quad\textbf{Discretization of $\boldsymbol{\alpha}$ and $\boldsymbol{\beta}$}

We start with discretizing $\boldsymbol{\alpha}$. To guarantee that only one operation dominates on each edge when the search process ends, we compute the following loss for each edge $\left(i,j\right)$:
\begin{equation}
\label{eq:entropy-alpha}
\mathcal{L}_{i,j}^\mathrm{O}(\boldsymbol{\alpha})=-{{\sum_{o\in\mathcal{O}}}\frac{\exp\left\{\alpha_{i,j}^o\right\}}{{\sum_{o'\in\mathcal{O}}}\exp\left\{\alpha_{i,j}^{o'}\right\}}\cdot \log\frac{\exp\left\{\alpha_{i,j}^o\right\}}{{\sum_{o'\in\mathcal{O}}}\exp\left\{\alpha_{i,j}^{o'}\right\}}}.
\end{equation}
Summarizing the loss term on all edges obtains the operation loss:
\begin{equation}
\mathcal{L}^\mathrm{O}(\boldsymbol{\alpha})=\sum_{j}\sum_{i<j}\mathcal{L}_{i,j}^\mathrm{O}(\boldsymbol{\alpha}).
\end{equation}
Note that Eq.~\eqref{eq:entropy-alpha} is an entropy-based loss function on $\frac{\exp\left\{\alpha_{i,j}^o\right\}}{{\sum_{o'\in\mathcal{O}}}\exp\left\{\alpha_{i,j}^{o'}\right\}}$, the probability of choosing $o\!\left(\cdot\right)$ as the operation of each $\left(i,j\right)$. Minimizing $\mathcal{L}^\mathrm{O}(\boldsymbol{\alpha})$ pushes the weights of all operations to a one-hot distribution, \textit{i.e.}, the probability of one operation is close to $1.0$ while that of others are close to $0.0$. Note that $\mathcal{L}^{O}(\boldsymbol{\alpha})$ is jointly optimized with $\mathcal{L}^{C}(\boldsymbol{\alpha},\boldsymbol{\theta})$, implying that when the search process is complete, the network parameters, $\boldsymbol{\theta}$, have been adjusted according to the one-hot $\boldsymbol{\alpha}$, consequently, the inaccuracy introduced by discretization is much smaller.

Things become a bit different when we try to discrete $\boldsymbol{\beta}$, because the configuration often requires to preserve more than one candidates, \textit{e.g.}, according to the standard DARTS formulation, each node receives input from two previous nodes. To handle it, we add an extra term to the previous entropy loss and constrain the maximum value of any $\beta_{i,j}$ to 1, and the overall loss is shown as: 
\begin{equation}
\label{eq:entropy-beta}
\mathcal{L}_{j}^\mathrm{E}(\boldsymbol{\beta})=-\sum_{i<j}{\frac{\exp\left\{\beta_{i,j}\right\}}{{\sum_{i'<j}\exp\left\{\beta_{i',j}\right\}}}\cdot   \log\frac{\exp\left\{\beta_{i,j}\right\}}{{\sum_{i'<j}\exp\left\{\beta_{i',j}\right\}}}}+\left|\sum_{B}{\beta_{i,j}}-2\right|^2,
\end{equation}
where $B = \{\beta_{i, j}| i < j, \beta_{i, j} > 0 \}$. Note that the sum of $2$ can be changed according to the search configuration. In the experimental part, we will show how this formulation generalizes to other types of desired topology, \textit{e.g.}, preserving $4$ or $6$ out of $14$ edges.

Similarly, summarizing this term on all nodes obtains the edge loss:
\begin{equation}
\mathcal{L}^\mathrm{E}(\boldsymbol{\beta})=\sum_{j}\mathcal{L}_{j}^\mathrm{E}(\boldsymbol{\beta}),
\end{equation}
and the discretization-aware objective function for architecture search is:
\begin{equation}
\label{eq:total_loss}
\mathcal{L}(\boldsymbol{\theta}, \boldsymbol{\alpha}, \boldsymbol{\beta}) = \mathcal{L}^\mathrm{C}(\boldsymbol{\alpha},\boldsymbol{\theta}) +\mathcal{L}^\mathrm{O}(\boldsymbol{\alpha})+\mathcal{L}^\mathrm{E}(\boldsymbol{\beta}),
\end{equation}

\noindent$\bullet$\quad\textbf{Discretization Priority}

Edge discretization and operation discretization depend on the performance estimation by each other. This warped paradox perplexes the community a lot for a long time, and can be eased by independently enforcing additional regularization on $\alpha$ and $\beta$. While, exploring the discretization priority of operation discretization and edge discretization further narrows the discretization gap. By introducing regularization control functions, the discretization-aware objective function for architecture search can be improved as:
\begin{equation}
\label{eq:total_loss}
\mathcal{L}(\boldsymbol{\theta}, \boldsymbol{\alpha}, \boldsymbol{\beta}) = \mathcal{L}^\mathrm{C}(\boldsymbol{\alpha}, \boldsymbol{\theta}) + \lambda_{c}(\lambda_{\boldsymbol{\alpha}}\mathcal{L}^\mathrm{O}(\boldsymbol{\alpha}) + \lambda_{\boldsymbol{\beta}}\mathcal{L}^\mathrm{E}(\boldsymbol{\beta})),
\end{equation}
where $\lambda_{c}$, $\lambda_{\boldsymbol{\alpha}}$ and $\lambda_{\boldsymbol{\beta}}$ are regularization control factors related to classification accuracy, operation discretization and edge discretization, respectively. 

Considering the dynamic change of node connections, operation weights, and network parameters during the searching process, the regularization factors are defined as functions of training epochs, and simplified to be chosen from five representative increasing functions, as shown in Figure~\ref{fig:dimension}, to reveal the regular pattern of optimization priority. At early training epochs, the network is not well trained. The regularization factors are small so that the training focuses on network parameters. As the optimization process continues, the network gets better trained and more attentions are paid on selecting operations and edges.

\subsection{Relationship to Prior Work}

There exist prior works to push the architectural parameters towards either $0$ or $1$ so as to align with the requirement of discretization.

For example, FairDARTS~\cite{chu2019fairdarts} introduced the zero-one loss as $-\frac{1}{K}\sum_{k=1}^{K}\left|\sigma\!\left(\alpha_{k}\right)-0.5\right|^{2}$ to quantize the architectural parameters, $\alpha$, by using individual sigmoid rather than softmax, where $\sigma\!\left(\cdot\right)$ indicates the sigmoid function. In addition, by considering NAS as an annealing process in which the system converges to a less chaotic status, XNAS~\cite{Nayman2019XNAS} proposed to reduce the temperature term of the cross-entropy loss so that weaker candidates get eliminated. However, FairDARTS was not able to control the exact number of preserved candidates -- sometimes there can be multiple weights pushing towards $1$ but only one of them is allowed to be kept; on the other hand, XNAS cannot support more than one candidates to be preserved, which suffers limited flexibility when applied to multi-choice scenarios. In comparison, our approach can adjust the loss function according to the desired topology -- we will show a variety of examples in Table~\ref{tab.imbalance}. If needed, it can freely generalize to choose multiple operations for each edge.

\section{Experiments}
\label{Experiments}
In this section, we first describe the experimental settings. We then validate the effect of our discretization-aware search approach. We also report the performance of our approach on balanced and imbalanced configurations, and compare it with the state-of-the-arts.

\subsection{Experimental Settings}
\textbf{Dataset} 
The commonly used CIFAR10 and ImageNet datasets are used to evaluate our network architecture search approach. CIFAR10 consists of 60K images, which are of a low spatial resolution of 32 $\times$ 32. The images are equally distributed over 10 classes, with 50K training and 10K testing images. 
ImageNet contains 1,000 object categories, which consists of 1.3M high-resolution training images and 50K validation images. The images are almost equally distributed overall classes. Following the commonly used settings, we apply the mobile setting where the input image size is ﬁxed to be 224 $\times$ 224 and the number of multi-add operations does not exceed 600M in the testing stage~\cite{xu2020pcdarts}. 

\textbf{Implementation Details.} 
Following DARTS as well as conventional architecture search approaches, we use an individual stage for architecture search, and after the optimal architecture is obtained, we use an additional process to train the classification model from scratch. During the search stage, the goal is to find the optimal \textbf{$\alpha$} and \textbf{$\beta$} under the entropy-based discretization regularization in an end-to-end manner. We search architectures on CIFAR10 and then transfer to ImageNet.

During the search procedure, we split the training data into two parts, one for each stage of the search process. As for search space, we follow DARTS but without zero as it requires to choose a low weight operation when zero has a advantage to form a standard cell. There are in total 7 options including 3$\times$3 and 5$\times$5 separable convolution, 3$\times$3 and 5$\times$5 dilated separable convolution, 3$\times$3 max-pooling, 3$\times$3 average-pooling, and skip-connect.

When searching, the over-parameterized super-network is constructed by stacking $8$ cells ($6$ normal cells and $2$ reduction cells) with the initial number of channels $16$, and each cell consists of N = $6$ nodes. The 50K training set of CIFAR10 is split into two subsets with equal size, with one subset used for training network weights and the other used for architecture hyper-parameters. 

We train the super-network for $50$ epochs and super-network weights are optimized by the momentum SGD algorithm, a momentum of $0.9$, and a weight decay of $3\times10^{-4}$.
The learning rate is reduced progressively to zero following a cosine schedule from an intial learning rate of $0.25$ without restart. We use an Adam optimizer~\cite{kingma2014adam} for both $\alpha$ and $\beta$, both with a ﬁxed learning rate of $3\times10^{-4}$, a momentum of $(0.5,0.999)$ and a weight decay of $10^{-3}$~\cite{xu2020pcdarts}. The memory cost of our implementation is smaller than $11$GB so that it can be trained on most modern GPUs.

\subsection{Results on CIFAR10}
\label{sec:effect}

In Table~\ref{tab.CIFAR10}, we compare the proposed approach with the state-of-the-art approaches. It can be seen that our approach outperforms the baseline DARTS method with a large margin (2.42\% vs. 2.76\%), and outperforms recent gradient-based methods including P-DARTS~\cite{chen2019progressive}, PC-DARTS~\cite{xu2020pcdarts},
and BayesNAS~\cite{zhou2019bayesnas}. Note that the significant performance gains are achieved with moderate parameter size (3.4 M) and computational cost (0.3 GPU days). The performance gains validate the effectiveness of our entropy-based regularization method and the importance of discretization-aware search itself. This part we will first introduce our approach to search standard cells (select $8$ edges from $14$ $i.e.$ balanced configuration), and then to further illustrate the effectiveness of our approach, we will search non-standard cells (imbalanced configuration).

\begin{table}[t]
\centering
\caption{Comparison of classification error (\%) with state-of-the-arts on CIFAR10. For DA\textsuperscript{2}S, $2.42\%$ and $2.51\%$ are the best and average errors, respectively, and the search cost, $0.3$ GPU-days, is reported on a single NVIDIA GTX-1080Ti GPU -- on a Tesla-V100 GPU, the time is expected to be $0.2$ GPU-days.}
\label{tab.CIFAR10}
\begin{threeparttable}[b]
\resizebox{0.95\textwidth}{!}{
\begin{tabular}{@{}lcccccc@{}}
\toprule
\multirow{2}{*}{\textbf{Architecture}} & \textbf{Test Err.} & \textbf{Params} & \textbf{Search Cost} & \multirow{2}{*}{\textbf{Search Method}} \\
&        \textbf{(\%)} & \textbf{(M)} & \textbf{(GPU-days)} &\\
\midrule
DenseNet-BC~\cite{huang2017densely}                       & 3.46  & 25.6 & -    & manual \\
\midrule
NASNet-A + cutout~\cite{zoph2018learning} & 2.65  & 3.3  & 1800 & RL      \\
AmoebaNet-B + cutout~\cite{real2018regularized} & 2.55$\pm$0.05 & 2.8  & 3150 & evolution \\
Hireachical Evolution~\cite{liu2017hierarchical} & 3.75$\pm$0.12 &  15.7 & 300  & evolution \\
PNAS~\cite{liu2018progressive}  & 3.41$\pm$0.09 & 3.2  & 225  & SMBO \\
ENAS + cutout~\cite{pham2018efficient}  & 2.89 & 4.6  & 0.5  & RL \\
NAONet-WS~\cite{luo2018neural}       & 3.53 & 3.1 & 0.4 & NAO\\
\midrule
DARTS (1st order) + cutout~\cite{liu2018darts} & 3.00$\pm$0.14 &  3.3 & 0.4  & gradient-based \\
DARTS (2nd order) + cutout~\cite{liu2018darts} & 2.76$\pm$0.09 &  3.3 & 1 & gradient-based \\
SNAS (moderate) + cutout~\cite{xie2018snas}    & 2.85$\pm$0.02 & 2.8  & 1.5  & gradient-based \\
ProxylessNAS + cutout~\cite{cai2018proxylessnas}   & 2.08 & -     &  4.0    & gradient-based \\
P-DARTS + cutout~\cite{chen2019progressive}                                    & 2.50 &  3.4  & 0.3 & gradient-based \\
PC-DARTS + cutout~\cite{xu2020pcdarts} & 2.57$\pm$0.07   & 3.6 &   \bf 0.1  & gradient-based \\
BayesNAS + cutout~\cite{zhou2019bayesnas}  & 2.81$\pm$0.04 &  3.4  & 0.2& gradient-based \\
\midrule
DA\textsuperscript{2}S (ours) + cutout  & \bf 2.42/2.51$\pm$0.09   & 3.4  & 0.3 & gradient-based \\
\bottomrule
\end{tabular}
}
\end{threeparttable}
\end{table}

\textbf{Operation and Edge Discretization}.
There are 7 operations in total for all cells (the `none' operation is not used).
Each cell has 14 edges and the network consists of two kinds of cells: the normal cells and the reduction cells, that the network architecture depends on the search of 28 edges. That is to say $\mathcal{L}^{O}(\alpha)$ is the sum of 28 operation entropy losses. 
And $\mathcal{L}^{E}(\beta)$ is the sum of all edge entropy losses, Eq.~\eqref{eq:entropy-beta}. 
In Eq.~\eqref{eq:total_loss}, we experimentally define that $\lambda_{\alpha} = \lambda_1$, and $\lambda_{\beta} = 4\lambda_2$. Then we evaluate the results with $\lambda_{1}$ and $\lambda_2$ under different setting of functions shown in the Figure~\ref{fig:dimension}.

\begin{figure}[t]
\centering
\includegraphics[width=1\columnwidth]{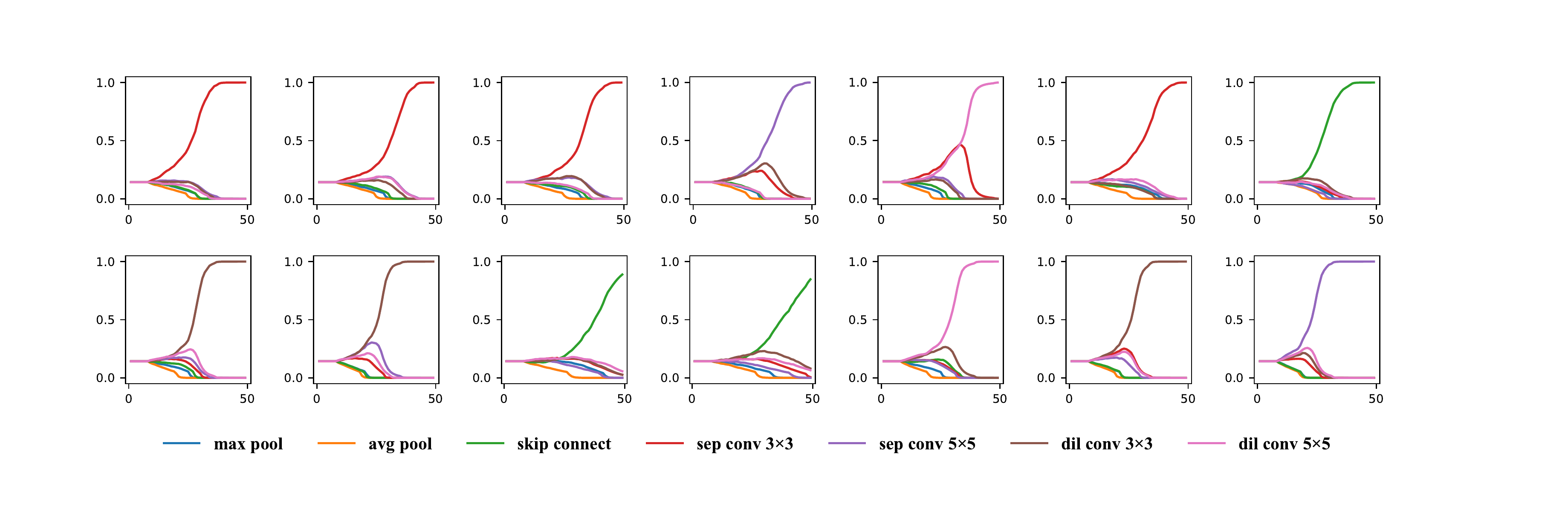}
\caption{Evolution of softmax of operation weights $\alpha$ during the searching procedure in a normal cell on CIFAR10. The horizontal axis denotes training epoch and vertical axis softmax weight value. (Best viewed in color with zooming in).}
\label{fig:alpha_evolution}
\end{figure}

\begin{figure}[t]
\centering
\includegraphics[width=0.8\columnwidth]{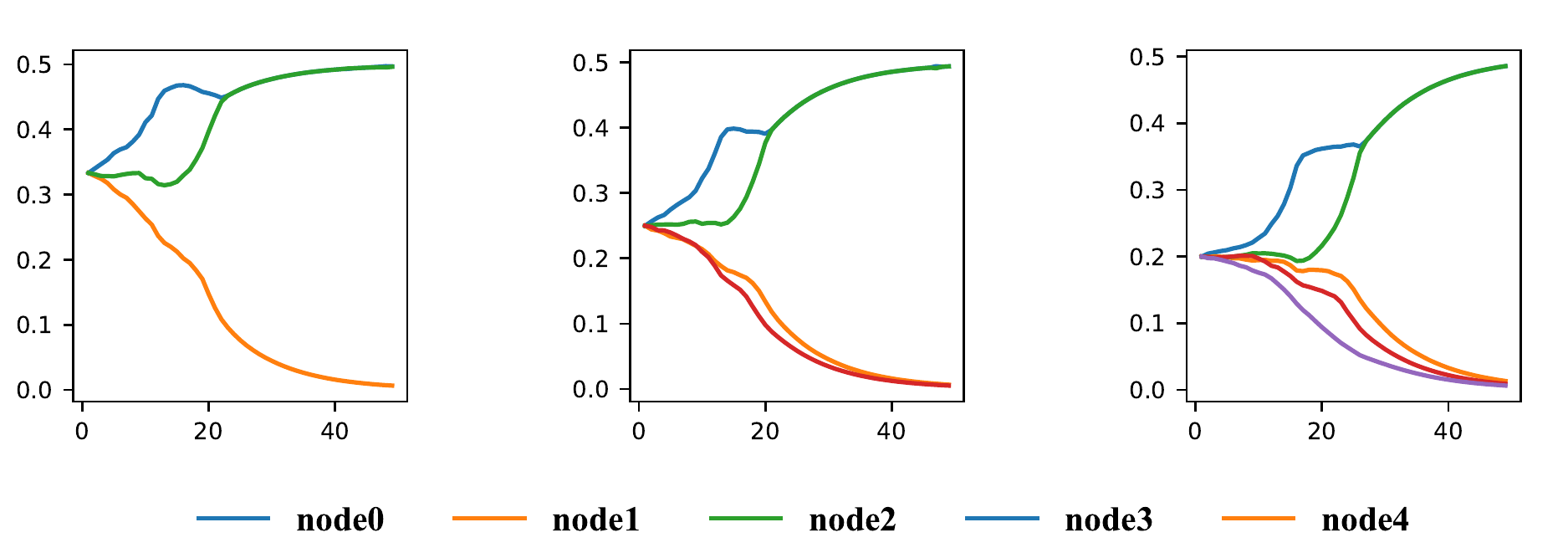}
\caption{Evolution of softmax of edge weights $\beta$ of node3/4/5 during the searching procedure in a normal cell searched on CIFAR10. The horizontal axis denotes training epoch and vertical axis softmax weight value.}
\label{fig:beta_evolution}
\end{figure}

In Figure~\ref{fig:alpha_evolution}, we present the evolution of softmax of operation weights $\alpha$ on CIFAR10 with $14$ edges in a normal cell. It can be seen that after about $20$ training epochs, the softmax of operation weights begin to significantly differentiate. At the final epoch, a single largest (towards 1) is obtained with the rest of small values (towards 0), which clearly demonstrate the effect of  operation discretization. In Figure~\ref{fig:beta_evolution}, we present the softmax  evolution of $\beta$, which validates the effect of edge discretization. Note that there are two edges selected at the same time for each pair of nodes, which shows the effectiveness of connection constraints in Eq.~\eqref{eq:entropy-beta}. 

\textbf{Discretization Priority.} 
The entropy loss function inevitably interferes the searching procedure of DARTS, particularly, at the early epochs. Therefore, we propose to progressively increase the regularization factors using monotonous functions as shown in Figure~\ref{fig:dimension}. In Table\ \ref{tab.func}, we fix $\lambda_{1}$ and $\lambda_{2}$ as `const' (equals $1.0$), it can be seen that the fast increasing functions, such as `linear' and `log', outperform slow ones for regularization factor $\lambda_{c}$, while `linear' achives the best performance. It can be explained that moderately quick (linear) enhancement of the regularization on classification loss may have the smallest interference to the searching procedure.

In Table\ \ref{tab:priority}, we test the priority of operation and edge discretization using different regularizaton control functions with $\lambda_{c}$ set to `linear' as default. We fix $\lambda_{1}$ as `const' and evaluate $\lambda_{2}$ using different control functions since the $8^{th}$ epochs before which $\lambda_{2}$ is fixed as 0. This means that the priority of operation discretization is higher than that of edge. Under this setting, the best performance (2.49\%) is achieved by the `step' function. On the other hand, we fix $\lambda_{2}=1.0$ and change $\lambda_{1}$ under the same conditions. The best performance (2.42\%) is achieved by the `log' function.

The higher performance obtained by fixing $\lambda_{2}=1.0$ shows that when the edge discretization dominates the search procedure, quick convergence of the topology of cell can lead the operation discretization-aware search converge better with fast increasing regularization control function (`log') utilized.

\begin{table}[t]
\centering
\caption{Classification errors ($\%$) under different control functions for regularization factor $\lambda_{c}$  on CIFAR10.}
\label{tab.func}
\begin{threeparttable}[b]
\setlength{\tabcolsep}{3mm}{
\begin{tabular}{@{}cccccc@{}}
\toprule
baseline& \textbf{const} & \textbf{log} & \textbf{exp} &\textbf{step}  &\textbf{linear} \\
\midrule
2.76$\pm$0.09 & 2.64$\pm$0.14   &2.56$\pm$0.06   & 2.78$\pm$0.11  & 2.60$\pm$0.07   &\bf 2.54$\pm$0.02\\

\bottomrule
\end{tabular}
}
\end{threeparttable}
\end{table}

\begin{flushleft}
    \begin{minipage}{\textwidth}
        \begin{minipage}{0.3\textwidth} 
            \flushleft
            \includegraphics[scale=0.27]{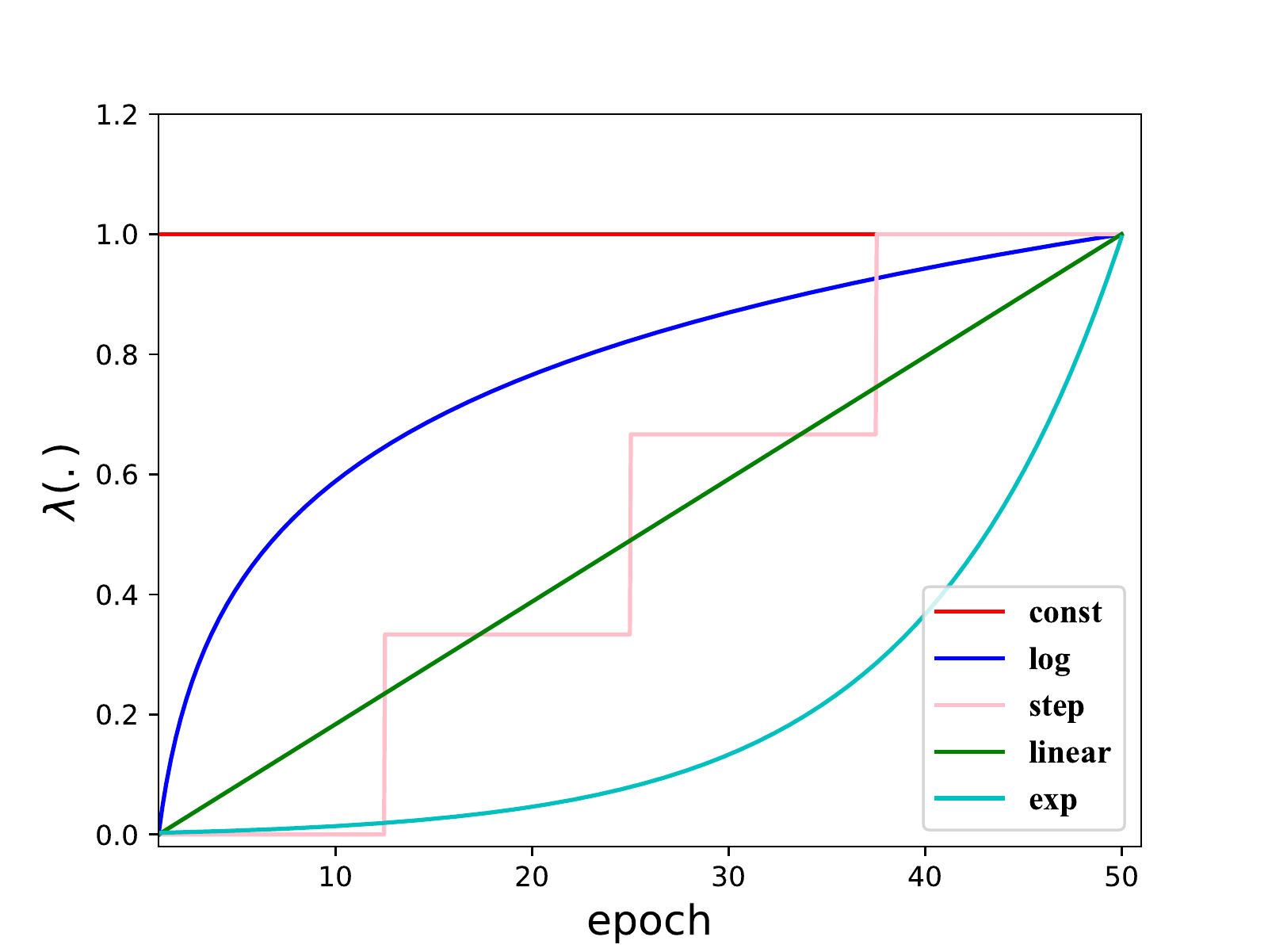}
            \captionof{figure}{Five regularization control functions. This figure is best viewed in color.} 
            \label{fig:dimension}
        \end{minipage} 
        \begin{minipage}{.7\textwidth} 
            \centering
            \begin{tabular}{ccccc}	
                \toprule
                \multirow{2}*{\textbf{Function}} & \multicolumn{2}{c}{\textbf{$\lambda_{1}$=1.0}} & \multicolumn{2}{c}{\textbf{$\lambda_{2}$=1.0}} \\
                & \textbf{best} & \textbf{average} & \textbf{best} & \textbf{average} \\
                \midrule
                const  &2.71  & 2.74$\pm$0.03  &2.53           &2.56$\pm$0.03 \\
                linear &2.55  & 2.57$\pm$0.02  &2.66           &2.68$\pm$0.02 \\
                exp   &2.51  & 2.53$\pm$0.02  &2.57           &2.60$\pm$0.03 \\
                step   &\bf 2.49  & 2.54$\pm$0.05  &2.51       &2.57$\pm$0.06 \\
                log   &2.61  & 2.64$\pm$0.03 &\bf{2.42}        &2.51$\pm$0.09 \\
                \bottomrule		
            \end{tabular}
            \captionof{table}{Classification errors ($\%$) when fixing either of $\lambda_{1}$ or $\lambda_{2}$ while changing the other using the regularization control functions on CIFAR10.} 
            \label{tab:priority}
        \end{minipage}
    \end{minipage}
\end{flushleft}

\textbf{Imbalanced Configurations.}
In the above settings, it is defined that there are two inputs for each node in cells and the optimization objective is to select 8 out of 14 edges. This constraint largely reduces the difficulty of search, $i.e.,$ a random search can find architectures of moderate accuracy. 
To further validate the effectiveness and generalization of our approach, we search architectures with imbalanced configurations. Specifically, we break the setting about choosing 8 from 14 and choosing fewer edges to magnify the gap between architectures before and after discretization.

Four configurations, namely, preserving $3$--$6$ out of $14$ edges, are used to validate our approach and compared it with DARTS. For DARTS, we use the default searched architecture and select 3, 4, 5, or 6 edges according to the weights of operations. For our approach, to select 3 edges, we pose edge entropy-loss on node2 and node3, and select the largest one, and pose edge entropy-loss on node4 and node5 to select one on each. To select 4 edges, we pose edge entropy-loss on four inner nodes so that each of them has a single edge. For 5 edge edges, we select two on node5 and one on other 3 nodes. For 6 edges, we select two on node4/node5 and one on node2/node3.

In Table~\ref{tab.imbalance}, the performance under imbalanced configurations of DARTS and our approach is compared. Under imbalanced configurations, the performance of DARTS dramatically drops in a large margin around [77.75-78.00], which demonstrates that the discretization process does bring a significant gap before and after prunning. Such gap has unpredictable impact upon searched architecture. In contrast, with discretization-aware constraint, our approach achieves relatively stable performance that the classification accuracy drop are significantly reduced to [0.21, 21.29]. For each configuration, it outperforms DARTS with significant margins (2.16\%, 1.75\%, 0.51\%, 0.27\%) after re-training.
\begin{table}[t]
\centering
\caption{Comparison (\%) of re-training error and super-network accuracy between DARTS and DA\textsuperscript{2}S under imbalanced configurations on CIFAR10. In the first column, $3/14$ indicates preserving $3$ out of $14$ edges.}
\label{tab.imbalance}
\begin{threeparttable}[b]
\setlength{\tabcolsep}{1.2mm}{
\begin{tabular}{@{}lcccccc@{}}
\toprule
\multirow{2}*{\textbf{config}} & \multicolumn{3}{c}{\textbf{DARTS~\cite{liu2018darts}}} & 
\multicolumn{3}{c}{\textbf{DA\textsuperscript{2}S}} 
\\
& \textbf{error} & \textbf{para(M)} & \textbf{acc. drop} 
& \textbf{error} & \textbf{para(M)} & \textbf{acc. drop} 
\\
\midrule
3/14
& 5.83$\pm$1.21 &1.5$\pm$0.2 &87.87$\rightarrow$10.03
&\bf 3.67$\pm$0.24 &1.9$\pm$0.2 &85.52$\rightarrow$64.23
\\
4/14
& 4.79$\pm$1.17 &1.9$\pm$0.3 &87.87$\rightarrow$09.87
&\bf 2.94$\pm$0.09  &2.5$\pm$0.1 &85.63$\rightarrow$85.42
\\
5/14
&3.23$\pm$0.08  &2.2$\pm$0.2 &87.87$\rightarrow$10.12
&\bf2.72$\pm$0.06   &2.9$\pm$0.1 &84.76$\rightarrow$71.85
\\
6/14
&2.91$\pm$0.05  &2.7$\pm$0.1 &87.87$\rightarrow$09.96
&\bf2.64$\pm$0.02   &3.0$\pm$0.1  &84.29$\rightarrow$64.24
\\

\bottomrule
\end{tabular}
}
\end{threeparttable}
\end{table}

\begin{figure}[t]
\centering
\begin{minipage}{0.49\textwidth}
\subfloat[the normal cell found on CIFAR10]{\includegraphics[width=1.\linewidth]{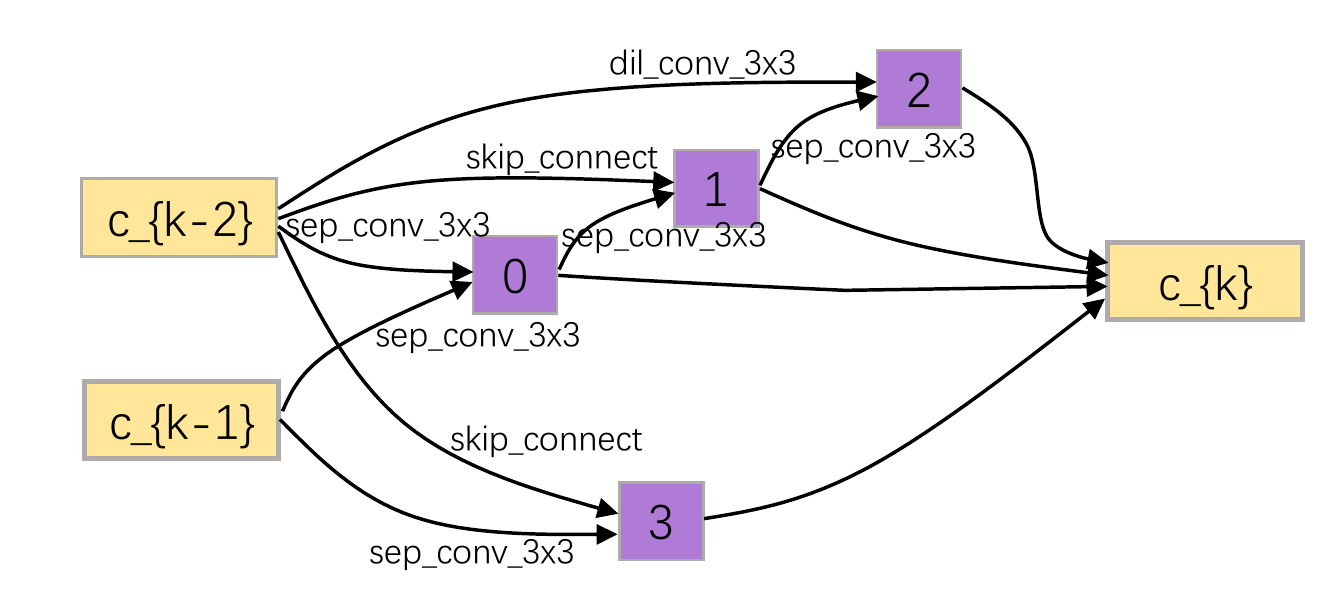}}\label{ncells_s1}\\
\end{minipage}
\begin{minipage}{0.49\textwidth}
\subfloat[the reduction cell found on CIFAR10]{\includegraphics[width=1.0\linewidth]{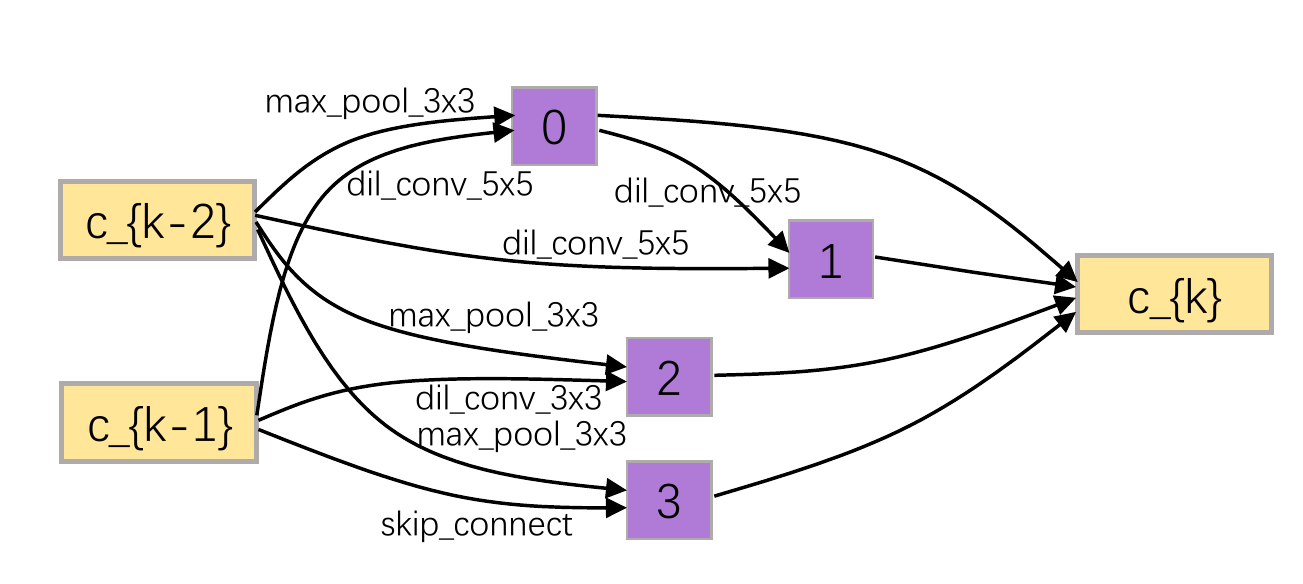}}\label{ncells_s3}\\
\end{minipage}
\caption{Normal cell and reduction cell searched on CIFAR10.}
\label{fig:cells}
\end{figure}

\subsection{Results on ImageNet}
This part we use large-scale ImageNet to test the transferability of cells searched on CIFAR10 as shown in Figure~\ref{fig:cells}. Same configuration as DARTS is adopt, $\textit{i.e.}$, the entire network is construct by stacking $14$ cells with an initial channel number of $48$. We train the network for $250$ epochs from scratch with batch size $1024$ on $8$ Tesla V100 GPUs. An SGD optimizer is used for optimizing the network parameters with an initial learning rate of $0.5$ (decayed linearly after each epoch), and also a momentum of 0.9 and a weight decay of $3\times10^{-5}$. Other enhancements including label smoothing~\cite{labelsmooth} and auxiliary loss are used during training, and learning rate warmup~\cite{warmup} is applied for the first $5$ epochs. 

In Table~\ref{tab.ImageNet}, we evaluate the proposed approach and compare the result with the state-of-the-art approaches under the mobile setting (the FLOPs does not exceed $600\mathrm{M}$). DA\textsuperscript{2}S outperforms the direct baseline, DARTS, by a significant margin of $2.3\%$ (an error rate of $24.4\%$ vs. $26.7\%$). DA\textsuperscript{2}S also produces competitive performance among some recently published work including P-DARTS, PC-DARTS, and BeyesNAS, when the network architecture is searched on CIFAR and transferred to ImageNet. This further verifies the superiority of our DA\textsuperscript{2}S in mitigating the discretization gap in the differentiable architecture search framework.

\begin{table}[t]
\centering
\caption{Comparison of classification error (\%) on ImageNet under the mobile setting (no larger than $600\mathrm{M}$ FLOPs).}
\label{tab.ImageNet}
\begin{threeparttable}[b]
\resizebox{0.95\textwidth}{!}{
\begin{tabular}{@{}lcccccc@{}}
\toprule
\multirow{2}{*}{\textbf{Architecture}} & \multicolumn{2}{c}{\textbf{Test Err. (\%)}} & \textbf{Params} & $\times+$ & \textbf{Search Cost} & \multirow{2}{*}{\textbf{Search Method}} \\
\cmidrule(lr){2-3}
&                            \textbf{top-1} & \textbf{top-5} & \textbf{(M)} & \textbf{(M)} & \textbf{(GPU-days)} &\\
\midrule
Inception-v1~\cite{szegedy2015going} & 30.2 & 10.1 & 6.6 & 1448 & -    & manual \\
MobileNet~\cite{howard2017mobilenets} & 29.4 & 10.5 & 4.2 & 569  & -    & manual \\
ShuffleNet 2$\times$ (v1)~\cite{zhang2018shufflenet} & 26.4 & 10.2 & $\sim$5  & 524  & -    & manual \\
ShuffleNet 2$\times$ (v2)~\cite{ma2018shufflenet} & 25.1 & - & $\sim$5  & 591  & -    & manual \\
\midrule
NASNet-A~\cite{zoph2018learning} & 26.0 & 8.4  & 5.3 & 564  & 1800 & RL \\
AmoebaNet-C~\cite{real2018regularized} & 24.3 & 7.6  & 6.4 & 570  & 3150 & evolution \\
PNAS~\cite{liu2018progressive} & 25.8 & 8.1  & 5.1 & 588  & 225  & SMBO \\
MnasNet-92~\cite{tan2018mnasnet} & 25.2 & 8.0  & 4.4 & 388  & -    & RL \\
\midrule
DARTS (2nd order)~\cite{liu2018darts}  & 26.7 & 8.7  & 4.7 & 574  & 4.0    & gradient-based \\
SNAS (mild)~\cite{xie2018snas}     & 27.3 & 9.2  & 4.3 & 522  & 1.5  & gradient-based \\
ProxylessNAS (GPU)~\cite{cai2018proxylessnas}  & 24.9 & 7.5  & 7.1 & 465  & 8.3  & gradient-based \\
P-DARTS (CIFAR10)~\cite{chen2019progressive} & 24.4 & 7.4  & 4.9 & 557  & 0.3  & gradient-based \\

PC-DARTS (CIFAR10)~\cite{xu2020pcdarts}   & 25.1  & 7.8  & 5.3   & 586  & 0.1  & gradient-based \\
BayesNAS~\cite{zhou2019bayesnas} & 26.5 & 8.9  & 3.9 & -  & 0.2  & gradient-based \\
\midrule

DA\textsuperscript{2}S (CIFAR10)   &\textbf{24.4}   & 7.3   & 5.0   &565   &0.3   & gradient-based\\

\bottomrule			
\end{tabular}
}
\end{threeparttable}
\end{table}


\section{Conclusions}

In this paper, we propose a discretization-aware NAS method, which works by introducing an entropy-based loss term to push the super-network towards a discretization-friendly status according to the pre-defined target. This strategy can be applied to either selecting an operator for each edge, or selecting a fixed number of edges for each node. Experiments on standard image classification benchmarks demonstrate the superiority of our approach, in particular, under some imbalanced configurations which were not studied before.

This work provides another evidence that one-shot neural architecture search can benefit from shrinking the gap between the super-network and sub-networks. As the search space becomes more complicated in the future, we expect our approach to serve as a standard tool to alleviate the discretization gap. We also look forward to investigate some uncovered problems, \textit{e.g.}, whether discretization can be done in a gradual manner so as to further reduce the error.

%
%
\bibliographystyle{splncs}
\bibliography{egbib}
\end{document}